\documentclass{article}
\usepackage{spconf,amsmath,graphicx, amssymb}

\usepackage{multirow}
\usepackage{multicol}
\usepackage{bm}

\title{Sequence training and adaptation of Highway Deep Neural Networks}
\name{Liang Lu \thanks{This work was done at The University of Edinburgh.}}
\address{Toyota Technological Institute at Chicago, USA  \\
{\small \tt llu@ttic.edu}}
\begin{document}
%\ninept
%
\maketitle
\begin{abstract}
Highway deep neural network (HDNN) is a type of depth-gated feedforward neural network, which has shown to be easier to train with more hidden layers and also generalise better compared to conventional plain deep neural networks (DNNs). Previously, we investigated a structured HDNN architecture for speech recognition, in which the two gate functions were tied across all the hidden layers, and we were able to train a much smaller model without sacrificing the recognition accuracy. In this paper, we carry on the study of this architecture with sequence-discriminative training criterion and speaker adaptation techniques on the AMI meeting speech recognition corpus. We show that these two techniques improve speech recognition accuracy on top of the model trained with the cross entropy criterion. Furthermore, we demonstrate that the two gate functions that are tied across all the hidden layers are able to control the information flow over the whole network, and we can achieve considerable improvements by only updating these gate functions in both sequence training and adaptation experiments. 

\end{abstract}
\section{Introduction}
\label{sec:intro}

Although they have been tremendously successful in the field of speech processing, neural network models are usually criticised to be lack of structure, less interpretable and less adaptable. Furthermore, most neural network acoustic models are much larger than conventional models using Gaussian mixtures, which make it challenging to deploy these models in resource constrained platforms such as embedded devices.  Recently, there have been some works to overcome these limitations. For example, Tan et al. in \cite{tan2015improving} investigated the stimulated learning of deep feedforward neural networks to make them more interpretable and to gain insight about the behaviour of those networks. On the other hand, in order to address the size of neural network acoustic models, small-footprint neural network models have received considerable research efforts such as using low-rank matrices~\cite{xue2013restructuring, sainath2013low},  teacher-student style training~\cite{li2014learning, ba2014deep, romero15_fitnet} and structured linear layers~\cite{le2013fastfood, sindhwani2015structured, moczulski2015acdc}. Small-footprint models are superior in several aspects. Apart from requiring lower computational cost and taking less memory, they may be more applicable for low-resource languages, where the amount of training data are usually much smaller. Furthermore, with smaller number of model parameters, these models may be more adaptable to the target domains, environments or speakers, when they are different from the training condition. 

Previously, we proposed a small-footprint acoustic model using highway deep neural network (HDNN)~\cite{llu2016a}. HDNN is a type of network with shortcut connections between hidden layers~\cite{srivastava2015training}. Compared to the plain networks with skip connections, HDNNs are equipped with two gate functions --  {\it transform} and {\it carry} gate -- to control and facilitate the information flow over all the whole network.  In particular, the transform gate is used to scale the output of a hidden layer and the carry gate is used to pass through the input directly after elementwise rescaling. The gate functions are the key to train very deep networks~\cite{srivastava2015training} and to speed up convergence as experimentally validated in~\cite{llu2016a}. Furthermore, for speech recognition, the recognition accuracy can be retained by simply increasing the depth of the network, while the  the number of hidden units in each hidden layer can be significantly reduced. As a result, the networks became much thinner and deeper with much smaller number of model parameters. In contrast to training plain feedforward neural networks of the same depth and width, we did not encounter any difficulty to train these highway networks using the standard stochastic gradient decent algorithm without pretraining~\cite{llu2016a}. %Overall, we were able to train much smaller highway networks without loosing the recognition accuracy. 

However, our study was focused on the cross entropy (CE) training of the networks previously, while in this paper, we investigate if our previous observations still hold in the case of sequence training. To further understand the effect of the gate functions in HDNNs, we performed the ablation experiments, in which we disabled the update of the model parameters in the hidden layers and/or classification layer during sequence training. Based on the experiments using the AMI meeting transcription corpus, we observed that by only updating the parameters in the gate functions, we were able to retain most of the improvement by sequence training, which supports our argument that the gate functions can manipulate the behaviour of all the hidden layers in the nonlinear feature extractor. Since the number of model parameters in the gate functions is relatively small, we then study the speaker adaptation techniques in the unsupervised fashion, in which we only fine tune the gate functions using the speaker dependent data. Using the seed models from both CE and sequence training, we were able to obtain consistent improvement by speaker adaptation. Overall, the small-footprint HDNN acoustic model with 5 million model parameters achieved slightly better results compared to the DNN baseline with 30 million parameters, while the HDNN model with 2 million parameters obtained only slightly lower accuracy compared to the baseline.

\section{Highway Deep Neural Networks}
\label{sec:hdnn}

To facilitate our discussion, we divide the parameters in a standard neural network acoustic model into two sets -- $\theta_c$ represents the model parameters in the classifier, and $\theta_h$ denotes the model parameters of the hidden layers in the feature extractor. Given an input acoustic frame ${\bm x}_t$ at the time step $t$, the feature extractor transforms the input into another feature representation as 
\begin{align}
\hat{\bm x}_t = f({\bm x}_t, \theta_h),
\end{align}
and the classifier predicts the label using a softmax function as
\begin{align}
 \hat{\bm y}_t = g(\hat{\bm x}_t, \theta_c).
\end{align}
In this paper, we focus on the feedforward neural network, in which $f(\cdot)$ is composed by multiple layers nonlinear transformations. 

Highway deep neural networks~\cite{srivastava2015training} augment the feature extractor with gate functions, in which the hidden layer may be represented as
\begin{align}
\label{eq:hw}
{\bm h}_l &= \sigma({\bm h}_{l-1}, \theta_l)\circ T({\bm h}_{l-1}, {\bm W}_T) \nonumber \\
& \qquad \quad + {\bm h}_{l-1}\circ C({\bm h}_{l-1}, {\bm W}_c)
\end{align}
where ${\bm h}_l$ denotes the hidden activations of $l$-th layer parameterised by $\theta_l$, and $\sigma$ denotes the activation function such as {\tt sigmoid} or {\tt tanh}; $T(\cdot)$ is the {\it transform} gate that scales the original hidden activations; $C(\cdot)$ is the {\it carry} gate, which scales the input before passing it directly to the next hidden layer; $\circ$ denotes elementwise multiplication; The outputs of $T(\cdot)$ and $C(\cdot)$ are constrained to be within $[0, 1]$, and we use the sigmoid function for both gates that are parameterised by $\mathbf{W}_T$ and $\mathbf{W}_c$ respectively. Following our previous work~\cite{srivastava2015training}, we tie the parameters in the gate functions across all the hidden layers, which can significantly save model parameters. In this work, we do not use any bias vector in the two gate functions. Since the parameters in $T(\cdot)$ and $C(\cdot)$ are layer-independent, we denote $\theta_g=(\bm W_T, \bm W_c)$, and we will look into the specific roles of these model parameters in sequence training and model adaptation experiments. 

Note that, although there are more computational steps for each hidden layer compared to plain DNNs due to the gate functions, the training speed can still be improved if the size of the weight matrix is smaller.  Furthermore, the matrices can be packed together as
\begin{align} 
\tilde{\bm W}_l = \left[ \bm W_l^\top, \bm W_T^\top, \bm W_c^\top \right]^\top,
\end{align}
where $\bm W_l^\top$ is the weight matrix in the $l$-th layer, and we then compute $\tilde{\bm W}_l\bm h_{l-1}$ once for all. By this trick, we can leverage on the power of GPUs on computing large matrix-matrix multiplications efficiently in the minibatch mode, which can speed up the training significantly. 

\subsection{Sequence Training}
\label{sec:smbr}

Our previous results of HDNNs are obtained with the CE training criterion, where the loss function is defined as 
\begin{align}
\mathcal{L}^{(CE)}(\theta) = - \sum_j y_{jt} \log \hat{y}_{jt}, 
\end{align}
where $j$ is the index of the hidden Markov model (HMM) state, and $\bm y_t$ denotes the ground truth label (the loss function is defined with one training utterance here for the simplicity of notation). However, state-of-the-art speech recognition systems are usually built with sequence-training techniques, where the loss function is defined as the sequence level. These approaches have been well understood for neural network acoustic models~\cite{kingsbury2009lattice, kingsbury2012scalable, Vesely:IS13, su2013error}. For instance, if we denote ${\bm X}$ as the sequence of acoustic frames ${\bm X} = \{\bm x_1, \ldots, \bm x_T\}$ and $\bm Y$ as the sequence of labels, where $T$ is the length of the signal, the loss function from the scalable minimum Bayesian risk criterion(sMBR)~\cite{gibson2006hypothesis, kingsbury2009lattice} is defined as
\begin{align}
\mathcal{L}^{(sMBR)}(\theta) = \frac{\sum_{\mathcal{W} \in \Phi}p({\bm X} \mid \mathcal{W})^k P(\mathcal{W})A(\bm Y, \hat{\bm Y})}{\sum_{\mathcal{W} \in \Phi}p({\bm X} \mid \mathcal{W})^k P(\mathcal{W})},
\end{align}
where $A(\bm Y, \hat{\bm Y})$ measures the state level distance between the ground truth and predicted labels; $\Phi$ denotes the hypothesis space represented by a† denominator lattice, and $\mathcal{W}$ is the word-level transcription; $k$ is the acoustic score scaling parameter. However, only applying the sequence training criterion without regularisation may lead to overfitting as observed in~\cite{ Vesely:IS13, su2013error}. To address this problem, we interpolate the sMBR loss function with the CE loss as used in~\cite{su2013error}:
\begin{align}
\label{eq:reg}
\mathcal{L}(\theta) = \mathcal{L}^{(sMBR)}(\theta) + p \mathcal{L}^{(CE)}(\theta), 
\end{align}
where $p \in \mathbb{R}^{+}$ is the smoothing parameter\footnote{The sequence training recipe used in this paper is adapted from the one developed by Y. Zhang et al. at the JSALT 2015 workshop~\cite{zhang2015highway}. }. In this paper, we only focus on the sMBR criterion since it can achieve comparable or slightly better results compared to the maximum mutual information (MMI) or minimum pone error (MPE) criterion~\cite{Vesely:IS13}. In the experimental section, we also study the effect of the regularisation term for different model parameter sets in the highway neural network acoustic models. 

\subsection{Adaptation}
\label{sec:spk}

Adaption of standard feedforward neural networks is challenging due to the large number of unstructured model parameters, while the amount of adaptation data is usually much smaller. Traditional approaches include input or output layer adaptation, while recently, researchers incorporate speaker dependent model parameters into the model space that can manipulate the behaviour or transform the output of the network. Techniques belong to this category may include speaker code~\cite{abdel2013fast}, LHUC~\cite{swietojanski2014learning} and multiple basis neural networks~\cite{wu2015multi}. The HDNN architecture studied in this paper is more structured in the sense that the parameters in the gate functions are layer-independent, and as will be demonstrated in the experimental section, they are able to control the behaviour of all the hidden layers. This motivates us to investigate the adaptation of highway gates by only fining tune these model parameters. Although the number of parameters in the gate functions are still much larger compared to the amount of adaptation data at the per-speaker level, the size of the gate functions is more controllable, as we can reduce the number of hidden units without sacrificing the accuracy by increasing the depth of the neural network~\cite{llu2016a}. Another adaptation approach is the input feature augmentation method such as using i-vectors~\cite{saon2013speaker}, which may be complimentary to our study, but it is not investigated in this paper.   

\begin{table}[t]
\caption{Comparison of DNN and HDNN system with CE and sMBR training. The DNN systems were built using Kaldi toolkit, where the networks were pre-trained using restricted Bolzman machines. Results are shown in terms of word error rates (WERs). We use $H$ to denote the size of hidden units, and $L$ the number of layers. $M$ indicates million model parameters.} \vskip 1mm
\label{tab:seq1}
\centering \small
\begin{tabular}{lc|cc|cc}
\hline 

\hline
  & & \multicolumn{2}{c|}{{\tt eval}} & \multicolumn{2}{c}{\tt dev}  \\
Model  & Size & CE & sMBR & CE & sMBR  \\ \hline
DNN-$H_{2048}L_{6}$ & $30 M$  & 26.8  & 24.6 & 26.0 & 24.3 \\
DNN-$H_{512}L_{10}$ & $4.6 M$ & 28.0 & 25.6 & 26.8 & 25.1\\
DNN-$H_{256}L_{10}$ & $1.7 M$ & 30.4 & 27.5 & 28.4 & 26.5\\ 
DNN-$H_{128}L_{10}$ & $0.71 M$ & 34.1 & 30.8 & 31.5 & 29.3 \\ \hline
HDNN-$H_{512}L_{10}$ & $5.1 M$ & 27.2 & 24.9 & 26.0 & 24.5 \\
HDNN-$H_{256}L_{10}$ & $1.8 M$ & 28.6 & 26.0 & 27.2 & 25.2 \\
HDNN-$H_{128}L_{10}$ & $0.74 M$ & 32.0 & 29.4 & 29.9 & 28.1 \\
HDNN-$H_{512}L_{15}$ & $6.4 M$ & 27.1 & 24.7 & 25.8 & 24.3 \\
HDNN-$H_{256}L_{15}$ & $2.1 M$ & 28.4 & 25.9 & 26.9 & 25.2 \\ \hline

 \hline
\end{tabular}
\vskip-4mm
\end{table}

\section{Experiments}
\label{sec:exp}

\subsection{System Setup}

Our experiments were performed on the individual headset microphone (IHM) subset of the AMI meeting speech transcription corpus~\cite{renals2007recognition}. The amount of training data is around 80 hours, corresponding to roughly 28 million frames. We used 40-dimensional fMLLR adapted features vectors normalised on the per-speaker level, which were then spliced by a context window of 15 frames (i.e. $\pm7$). The number of tied HMM states is 3927. The HDNN models were trained using the CNTK toolkit~\cite{yu2014introduction}, while the results were obtained using the Kaldi decoder~\cite{povey2011kaldi}. We also used the Kaldi toolkit to compute the alignment and lattices for sequence training, as well as feature transforms. We set the momentum to be 0.9 after the 1st epoch for CE training, and we used the sigmoid activation for all the networks. The weights in each hidden layer of HDNNs were randomly initialised with a uniform distribution in the range of $[-0.5, 0.5]$ and the bias parameters were initialised to be $0$. We used a trigram language model for decoding. In order to make the experimental results comparable, we used the same training and cross-validation sets split and alignment for both HDNN and plain DNN systems. All the systems used the same decision tree for state tying. 

\begin{table}[t]
\caption{Results of switching off the update of different model parameters in sequence training. $\theta_h$ denotes all the model parameters of the hidden layers, $\theta_g$ denotes the parameters in the two gate functions, and $\theta_c$ is the parameters in the softmax layer.} \vskip 1mm
\label{tab:seq2}
\centering \small
\begin{tabular}{l|ccc|c}
\hline 

\hline
& \multicolumn{3}{c|}{sMBR Update}  &  WER \\
Model    & $\theta_h$ & $\theta_g$ & $\theta_c$  & (\tt eval) \\ \hline
 & $\times$ & $\times$ & $\times$ & 27.2 \\ 
HDNN-$H_{512}L_{10}$ & $\surd$ & $\surd$ & $\surd$ & 24.9 \\ 
  &  $\times$ & $\surd$ & $\surd$ & 25.2\\ 
 & $\times$ & $\surd$ & $\times$ & 25.8\\ \hline
  & $\times$ & $\times$ & $\times$ & 28.6 \\ 
HDNN-$H_{256}L_{10}$  &  $\surd$ & $\surd$ & $\surd$ & 26.0\\ 
  & $\times$ & $\surd$ & $\surd$ & 26.6 \\ 
 & $\times$ & $\surd$ & $\times$ & 27.0 \\ \hline
  & $\times$ & $\times$ & $\times$ & 27.1 \\ 
HDNN-$H_{512}L_{15}$  &  $\surd$ & $\surd$ & $\surd$ & 24.7 \\ 
  & $\times$ & $\surd$ & $\surd$ & 25.2 \\ 
 & $\times$ & $\surd$ & $\times$ & 25.6 \\ \hline
   & $\times$ & $\times$ & $\times$ & 28.4 \\ 
HDNN-$H_{256}L_{15}$  &  $\surd$ & $\surd$ & $\surd$ & 25.9 \\ 
  & $\times$ & $\surd$ & $\surd$ & 26.4 \\ 
 & $\times$ & $\surd$ & $\times$ & 26.6 \\ \hline
  
 \hline
\end{tabular}
\vskip-4mm
\end{table}

\begin{figure*}[t]
\small
\centerline{\includegraphics[width=0.73\textwidth]{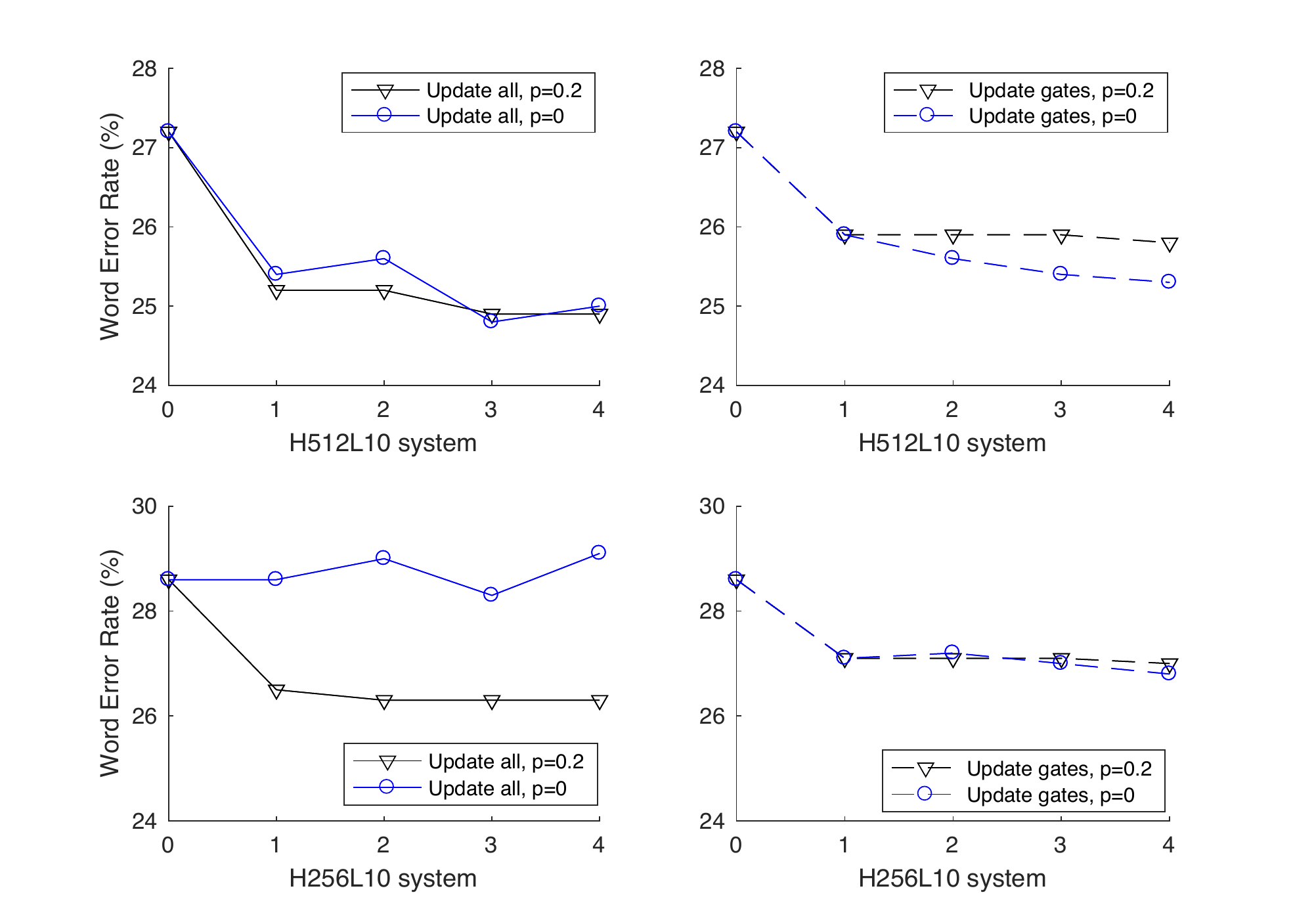}} \vskip-3mm
\caption{Convergence curves of sMBR training with and without the CE regularisation. The regularisation term can stabilise the convergence when updating all the model parameter, while its role is diminishing when updating the gate functions only.}  \vskip-3mm
\label{fig:reg}
\vskip-3mm
\end{figure*}

\subsection{Sequence Training}

In~\cite{llu2016a}, we showed that a smaller HDNN acoustic model was comparable to a much larger plain DNN model in terms of the accuracy when both were trained with the CE criterion, and it performed much better compared to DNNs of similar size. In this experiment, we investigate if this observation still holds after sequence training. We performed the sMBR update for 4 iterations, and we set the learning rate per frame to be $1\times 10^{-5}$ for both CNTK and Kaldi systems. We set the regularisation parameter $p$ to be 0.2 in Eq. \eqref{eq:reg} to avoid overfitting for CNTK systems, while we followed the recipe in~\cite{Vesely:IS13} for Kaldi systems using a slightly different regularisation technique. The baseline DNN systems were reproduced using the up-to-date Kaldi toolkit, and we obtained slightly better results compared to those in~\cite{llu2016a}. We did not use the CNTK to train plain DNN models because it did not converge with random initialisation for thin and deep networks as reported previously~\cite{llu2016a}. Table \ref{tab:seq1} shows the sequence training results of the plain DNN and HDNN systems, from which, we see that sequence training improves the recognition accuracy comparably for both DNN and HDNN systems, and the improvements are consistent for both {\tt eval}  and {\tt dev} sets as shown by Table \ref{tab:seq1}. Again, the HDNN model with around 5 million model parameters is on par with the plain DNN system with 30 million model parameters in terms of the recognition accuracy. In what follows, we only present results of the {\tt eval} set.

\begin{table}[t]
\caption{Results of sMBR training with and without regularisation. } \vskip 1mm
\label{tab:smbr2}
\centering \small
\begin{tabular}{l|c|cc}
\hline 

\hline
& &  \multicolumn{2}{c}{WER ({\tt eval})}   \\
Model   & sMBR Update   & $p=0.2$ & $p=0$   \\ \hline
HDNN-$H_{512}L_{10}$& $\{\theta_h, \theta_g, \theta_c \}$  & 24.9 & 25.0  \\
HDNN-$H_{512}L_{10}$&  $\theta_g$ & 25.8 & 25.3  \\
HDNN-$H_{256}L_{10}$ & $\{\theta_h, \theta_g, \theta_c \}$  & 26.0 &  29.1 \\ 
HDNN-$H_{256}L_{10}$&  $\theta_g$ & 27.0 & 26.8  \\ \hline

 \hline
\end{tabular}
\vskip-5mm
\end{table}

In the previous experiments, we updated all the model parameters in the HDNNs during sequence training. To look into the effect of a specific parameter set, we performed a set of ablation experiments, in which we switched off the update of some model parameters. These results are given in Table \ref{tab:seq2}, which show that only updating the parameters in the gates $\theta_g$ can retain most of the improvement given by sequence training, while updating $\theta_g$ and $\theta_c$ is close to optimum. Note that, $\theta_g$ only accounts for a small fraction of the total number of parameters, e.g., $\sim$ 10\% for the HDNN-$H_{512}L_{10}$ system and $\sim$ 7\% for the HDNN-$H_{256}L_{10}$ system. However, these results demonstrate that the gate functions can largely manipulate the behaviour of the neural network feature extractor. 

We then investigated the effect of the regularisation term in Eq. \eqref{eq:reg} for sequence training. We performed the experiments with and without the CE regularisation for two system settings, i.e., i) update all the model parameters; ii) update only the gate functions. Our motivation is to validate if only updating the gate parameters is more resistant to overfitting. The results are given in Table \ref{tab:smbr2}, from which we see that by switching off the CE regularisation term, we can achieve even slightly lower WER when updating the gate functions only. However, when updating all the model parameters, the regularisation term turned to be an important stabiliser for convergence. Figure \ref{fig:reg} shows the convergence curves for the two system settings. Overall, while the gate functions can largely control the behaviour of the highway networks, they are not prone to overfitting when other model parameters are switched off for update. 

\subsection{Unsupervised Adaptation}

The observations in the sequence training experiments inspired us to study the speaker adaptation of the gate functions, because they can control the behaviour the feature extractor with relatively small number of model parameters. In this paper, we use the term of {\it speaker adaptation} as convention, though the {\it speaker} can be defined as a cluster of acoustic frames at any granularity. We firstly performed the experiments in the unsupervised speaker adaptation setting, in which we decoded the evaluation set using the speaker-independent models, and then used the pseudo labels to fine tune the parameters in $\theta_g$ in the second pass. The evaluation set has around 8.6 hours of audio, and the number of speakers is 63. On average, each speaker has around 8 minutes speech, which corresponds to about 50 thousand frames. Compared to the size of $\theta_g$ in HDNNs, the amount of the adaptation data is still small, e.g., the size of $\theta_g$ in the HDNN-$H_{512}L_{10}$ system is around 0.5 million. We set the learning rate to be $2\times 10^{-4}$ per sample, and we updated $\theta_g$ for 5 iterations. 

\begin{figure}[t]
\small
\centerline{\includegraphics[width=0.465\textwidth]{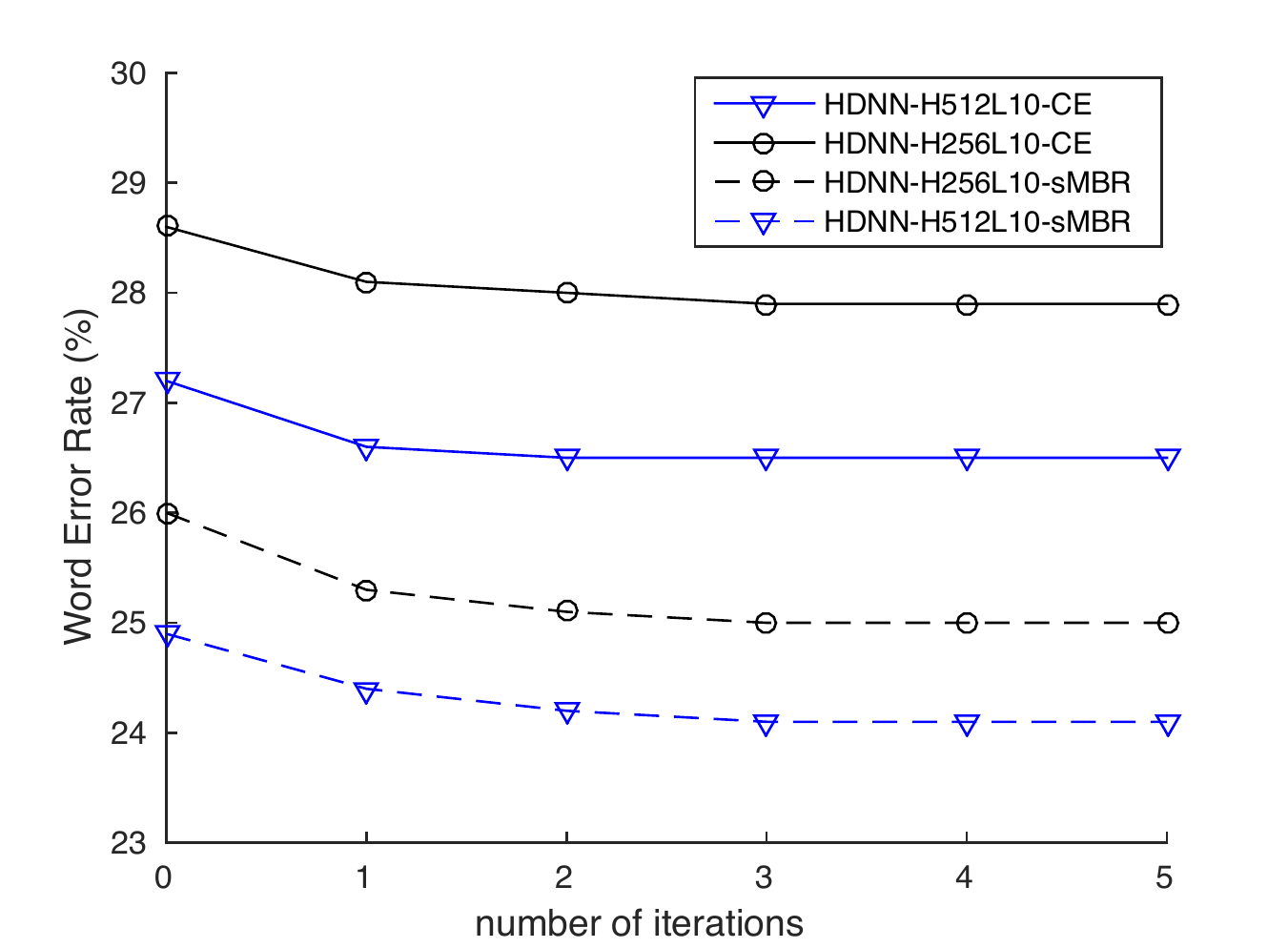}} 
\caption{Unsupervised adaptation results with different number of adaptation iterations. The speaker-independent models were trained by CE or sMBR, and we used CE criterion for all adaptation experiments. }  
\label{fig:adapt}
\vskip-3mm
\end{figure}

Table \ref{tab:adapt} shows the adaptation results, from which we observe small but consistent WER reduction with different model configurations on top of the speaker adapted features using fMLLR. Notably, the improvements are consistent for both seed models, where the speaker-independent models are trained using either the CE or the sMBR criterion. Updating all the model parameters yields smaller improvements as shown by the table. With speaker adaptation and sequence training, the HDNN system with 5 million model parameters (HDNN-$H_{512}L_{10}$) works slightly better than the DNN baseline with 30 million parameters (24.1\% vs. 24.6\%), while the HDNN model with 2 million parameters (HDNN-$H_{256}L_{10}$) achieves only slightly higher WER compared to the baseline (25\% vs. 24.6\%). In Figure \ref{fig:adapt} we show the adaptation results with different number of iterations. We observe that the best results can be achieved by only 2 or 3 adaptation iterations, thought updating the gate functions $\theta_g$ further does not yield overfitting. To further validate this, we also did experiments with 10 adaptation iterations, but we still did not observe overfitting. This observation is in line with the that in the sequence training experiments, demonstrating that the gate functions are relatively resistant to overfitting.

\begin{figure}[t]
\small
\centerline{\includegraphics[width=0.465\textwidth]{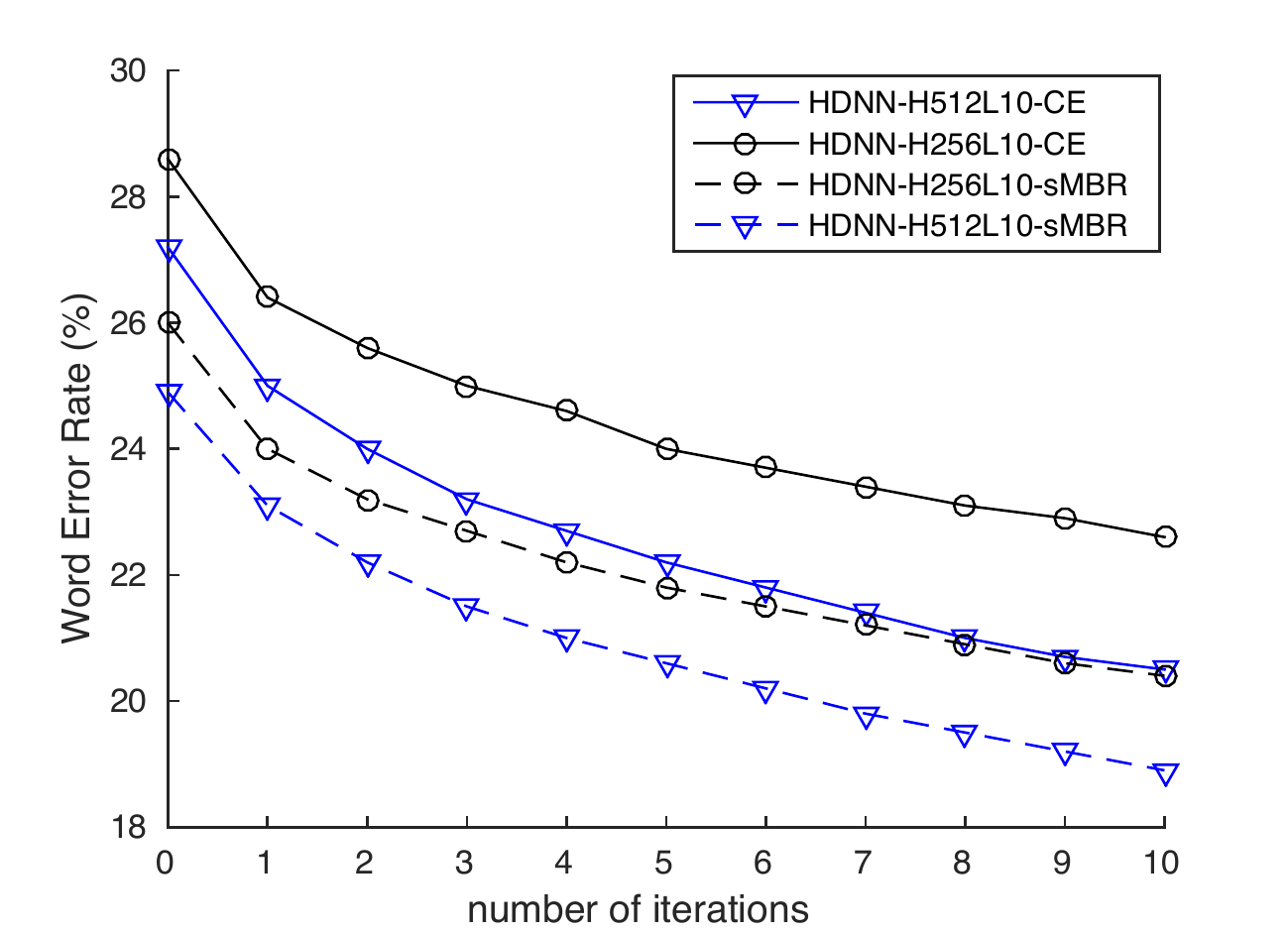}} 
\caption{Supervised adaptation results with oracle labels.}  
\label{fig:oracle}
\vskip-3mm
\end{figure}

\begin{table}[t]
\caption{Results of unsupervised speaker adaptation. Here, we only updated $\theta_g$ using the CE criterion, while the speaker-independent (SI) models were trained by either CE or sMBR. SD denotes speaker-dependent models. } \vskip 1mm
\label{tab:adapt}
\centering \small
\begin{tabular}{l|c|c|cc}
\hline 

\hline
& & & \multicolumn{2}{c}{WER ({\tt eval})}   \\
Model   & Seed   & Update & SI & SD   \\ \hline
HDNN-$H_{512}L_{10}$&   & & 27.2 & 26.5  \\
HDNN-$H_{256}L_{10}$ & CE  & & 28.6 & 27.9 \\
HDNN-$H_{512}L_{15}$ &   & & 27.1 & 26.4  \\ 
HDNN-$H_{256}L_{15}$ &   & $\theta_g$ & 28.4 &  27.6 \\ \cline{1-2}\cline{4-5}

HDNN-$H_{512}L_{10}$ &   & & 24.9 & {\bf 24.1} \\
HDNN-$H_{256}L_{10}$ &  & & 26.0 & {\bf 25.0}  \\
HDNN-$H_{512}L_{15}$ &  sMBR & & 24.7 &   24.0 \\ 
HDNN-$H_{256}L_{15}$ &   &  & 25.9 & 24.9 \\ \cline{3-5}
HDNN-$H_{512}L_{10}$ &   & $\{\theta_h, \theta_g, \theta_c\}$ & 24.9 & 24.5 \\
HDNN-$H_{256}L_{10}$ &  & & 26.0 & 25.4  \\ \hline

 \hline
\end{tabular}
\vskip-4mm
\end{table}

\subsection{Adaptation with Oracle Labels }

In the previous experiments, we studied the unsupervised adaptation condition, in which we obtained the labels for adaptation from the first-pass decoding.  In order to evaluate the impact of the accuracy of the labels to this adaptation method, we performed a set of diagnostic experiments, in which we used the oracle labels for adaptation. We obtained the oracle labels from the force alignment using the DNN model trained with the CE criterion and word level transcriptions. We have also fixed the alignment for all the adaptation experiments in order to compare the results from different seed models. Figure \ref{fig:oracle} shows the adaptation results with oracle labels, which demonstrates that significant WER reduction can be achieved when the supervision labels are more accurate. Therefore, the gate functions may have large capacity for adaptation with high quality pseudo labels. To further study this aspect, in the future, we shall investigate supervised adaptation of highway networks.

\section{Conclusions}

Highway deep neural networks are structured, depth-gated feedforward neural networks. In this paper, we studied sequence training and adaptation of these networks for acoustic modelling. In particular, we investigated the roles of the parameters in the hidden layers, gate functions and classification layer in the case of sequence training. We show that the gate functions, which only accounts for a small fraction of the whole parameter set, are able to control the information flow and adjust the behaviour of the neural network feature extractors. We demonstrated this in both sequence training and adaptation experiments, in which, considerable improvements were achieved by only updating the gate functions. By this two techniques, we obtained comparable or slightly lower WERs with much smaller acoustic models compared to a strong baseline set by the conventional DNN acoustic model with sequence training. Since the number of model parameters are still relative large compared to the speaker-level adaptation data, this adaptation technique may be more applicable in the domain adaptation scenarios, where the amount of adaptation data is relatively large. In the future, we shall also investigate the model compression techniques to further improve the results of our small-footprint acoustic models.

%\newpage

\small
\bibliographystyle{IEEEtran}
\bibliography{bibtex}

\begin{thebibliography}{10}
\providecommand{\url}[1]{#1}
\csname url@samestyle\endcsname
\providecommand{\newblock}{\relax}
\providecommand{\bibinfo}[2]{#2}
\providecommand{\BIBentrySTDinterwordspacing}{\spaceskip=0pt\relax}
\providecommand{\BIBentryALTinterwordstretchfactor}{4}
\providecommand{\BIBentryALTinterwordspacing}{\spaceskip=\fontdimen2\font plus
\BIBentryALTinterwordstretchfactor\fontdimen3\font minus
  \fontdimen4\font\relax}
\providecommand{\BIBforeignlanguage}[2]{{%
\expandafter\ifx\csname l@#1\endcsname\relax
\typeout{** WARNING: IEEEtran.bst: No hyphenation pattern has been}%
\typeout{** loaded for the language `#1'. Using the pattern for}%
\typeout{** the default language instead.}%
\else
\language=\csname l@#1\endcsname
\fi
#2}}
\providecommand{\BIBdecl}{\relax}
\BIBdecl

\bibitem{tan2015improving}
S.~Tan, K.~C. Sim, and M.~Gales, ``Improving the interpretability of deep
  neural networks with stimulated learning,'' in \emph{Proc. ASRU}.\hskip 1em
  plus 0.5em minus 0.4em\relax IEEE, 2015, pp. 617--623.

\bibitem{xue2013restructuring}
J.~Xue, J.~Li, and Y.~Gong, ``Restructuring of deep neural network acoustic
  models with singular value decomposition.'' in \emph{Proc. INTERSPEECH},
  2013, pp. 2365--2369.

\bibitem{sainath2013low}
T.~N. Sainath, B.~Kingsbury, V.~Sindhwani, E.~Arisoy, and B.~Ramabhadran,
  ``Low-rank matrix factorization for deep neural network training with
  high-dimensional output targets,'' in \emph{Proc. ICASSP}.\hskip 1em plus
  0.5em minus 0.4em\relax IEEE, 2013, pp. 6655--6659.

\bibitem{li2014learning}
J.~Li, R.~Zhao, J.-T. Huang, and Y.~Gong, ``Learning small-size dnn with
  output-distribution-based criteria,'' in \emph{Proc. INTERSPEECH}, 2014.

\bibitem{ba2014deep}
J.~Ba and R.~Caruana, ``Do deep nets really need to be deep?'' in \emph{Proc.
  NIPS}, 2014, pp. 2654--2662.

\bibitem{romero15_fitnet}
R.~Adriana, B.~Nicolas, K.~Samira~Ebrahimi, C.~Antoine, G.~Carlo, and
  B.~Yoshua, ``{Fitnets: Hints for thin deep nets},'' in \emph{Proc. ICLR},
  2015.

\bibitem{le2013fastfood}
Q.~Le, T.~Sarl{\'o}s, and A.~Smola, ``Fastfood-approximating kernel expansions
  in loglinear time,'' in \emph{Proc. ICML}, 2013.

\bibitem{sindhwani2015structured}
V.~Sindhwani, T.~N. Sainath, and S.~Kumar, ``Structured transforms for
  small-footprint deep learning,'' in \emph{Proc. NIPS}, 2015.

\bibitem{moczulski2015acdc}
M.~Moczulski, M.~Denil, J.~Appleyard, and N.~de~Freitas, ``{ACDC: A Structured
  Efficient Linear Layer},'' in \emph{Proc. ICLR}, 2016.

\bibitem{llu2016a}
L.~Lu and S.~Renals, ``Small-footprint deep neural networks with highway
  connections for speech recognition,'' in \emph{Proc. INTERSPEECH}, 2016.
  [Online] http://arxiv.org/pdf/1512.04280v3.pdf.

\bibitem{srivastava2015training}
R.~K. Srivastava, K.~Greff, and J.~Schmidhuber, ``Training very deep
  networks,'' in \emph{Proc. NIPS}, 2015.

\bibitem{kingsbury2009lattice}
B.~Kingsbury, ``Lattice-based optimization of sequence classification criteria
  for neural-network acoustic modeling,'' in \emph{Proc. ICASSP}.\hskip 1em
  plus 0.5em minus 0.4em\relax IEEE, 2009, pp. 3761--3764.

\bibitem{kingsbury2012scalable}
B.~Kingsbury, T.~N. Sainath, and H.~Soltau, ``Scalable minimum bayes risk
  training of deep neural network acoustic models using distributed
  hessian-free optimization,'' in \emph{Proc. INTERSPEECH}, 2012.

\bibitem{Vesely:IS13}
K.~Vesel\'{y}, A.~Ghoshal, L.~Burget, and D.~Povey, ``Sequence-discriminative
  training of deep neural networks,'' in \emph{Proc. INTERSPEECH}, 2013.

\bibitem{su2013error}
H.~Su, G.~Li, D.~Yu, and F.~Seide, ``Error back propagation for sequence
  training of context-dependent deep networks for conversational speech
  transcription,'' in \emph{Proc. ICASSP}.\hskip 1em plus 0.5em minus
  0.4em\relax IEEE, 2013, pp. 6664--6668.

\bibitem{gibson2006hypothesis}
M.~Gibson and T.~Hain, ``Hypothesis spaces for minimum bayes risk training in
  large vocabulary speech recognition.'' in \emph{Proc. INTERSPEECH}.\hskip 1em
  plus 0.5em minus 0.4em\relax Citeseer, 2006.

\bibitem{zhang2015highway}
Y.~Zhang, G.~Chen, D.~Yu, K.~Yao, S.~Khudanpur, and J.~Glass, ``{Highway Long
  Short-Term Memory RNNs for Distant Speech Recognition},'' \emph{Proc.
  ICASSP}, 2015.

\bibitem{abdel2013fast}
O.~Abdel-Hamid and H.~Jiang, ``Fast speaker adaptation of hybrid nn/hmm model
  for speech recognition based on discriminative learning of speaker code,'' in
  \emph{Proc. ICASSP}.\hskip 1em plus 0.5em minus 0.4em\relax IEEE, 2013, pp.
  7942--7946.

\bibitem{swietojanski2014learning}
P.~Swietojanski and S.~Renals, ``Learning hidden unit contributions for
  unsupervised speaker adaptation of neural network acoustic models,'' in
  \emph{Proc. SLT}.\hskip 1em plus 0.5em minus 0.4em\relax IEEE, 2014, pp.
  171--176.

\bibitem{wu2015multi}
C.~Wu and M.~J. Gales, ``Multi-basis adaptive neural network for rapid
  adaptation in speech recognition,'' in \emph{Proc. ICASSP}.\hskip 1em plus
  0.5em minus 0.4em\relax IEEE, 2015, pp. 4315--4319.

\bibitem{saon2013speaker}
G.~Saon, H.~Soltau, D.~Nahamoo, and M.~Picheny, ``Speaker adaptation of neural
  network acoustic models using i-vectors.'' in \emph{Proc. ASRU}, 2013, pp.
  55--59.

\bibitem{renals2007recognition}
S.~Renals, T.~Hain, and H.~Bourlard, ``{Recognition and understanding of
  meetings the AMI and AMIDA projects},'' in \emph{Proc. ASRU}.\hskip 1em plus
  0.5em minus 0.4em\relax IEEE, 2007, pp. 238--247.

\bibitem{yu2014introduction}
D.~Yu, A.~Eversole, M.~Seltzer, K.~Yao, Z.~Huang, B.~Guenter, O.~Kuchaiev,
  Y.~Zhang, F.~Seide, H.~Wang \emph{et~al.}, ``An introduction to computational
  networks and the computational network toolkit,'' Tech. Rep. MSR, Microsoft
  Research, Tech. Rep., 2014.

\bibitem{povey2011kaldi}
D.~Povey, A.~Ghoshal, G.~Boulianne, L.~Burget, O.~Glembek, N.~Goel,
  M.~Hannemann, P.~Motl{\i}cek, Y.~Qian, P.~Schwarz, J.~Silovsk\'y, G.~Semmer,
  and K.~Vesel\'y, ``{The Kaldi speech recognition toolkit},'' in \emph{Proc.
  ASRU}, 2011.

\end{thebibliography}

\end{document}